\documentclass[
]{ceurart}

\usepackage{listings}
\usepackage{adjustbox}
\usepackage{algorithm}
\usepackage{algorithmic}
\usepackage{subcaption}
\usepackage{float}
\usepackage{booktabs}
\usepackage{multirow}
\usepackage{makecell}
\usepackage{xcolor}

\begin{document}




\title{A Markovian Framing of WaveFunctionCollapse\\ for Procedurally Generating Aesthetically Complex Environments}

\author[1]{Franklin Yiu}[email=fyy2003@nyu.edu]
\author[1]{Mohan Lu}[email=ml7612@nyu.edu]
\author[1]{Nina Li}[email=nl2539@nyu.edu]
\author[1]{Kevin Joseph}[email=kj2676@nyu.edu]
\author[1]{Tianxu Zhang}[email=tz2902@nyu.edu]
\author[1]{Julian Togelius}[email=julian@togelius.com]
\author[1]{Timothy Merino}[email=tm3477@nyu.edu]
\author[1]{Sam Earle}[email=sam.earle@nyu.edu]

\address[1]{New York University, United States}

\begin{abstract}
Procedural content generation often requires satisfying both designer-specified objectives and adjacency constraints implicitly imposed by the underlying tile set. To address the challenges of jointly optimizing both constraints and objectives, we reformulate WaveFunctionCollapse (WFC) as a Markov Decision Process (MDP), enabling external optimization algorithms to focus exclusively on objective maximization while leveraging WFC’s propagation mechanism to enforce constraint satisfaction. We empirically compare optimizing this MDP to traditional evolutionary approaches that jointly optimize global metrics and local tile placement. Across multiple domains with various difficulties, we find that joint optimization not only struggles as task complexity increases, but consistently underperforms relative to optimization over the WFC-MDP, underscoring the advantages of decoupling local constraint satisfaction from global objective optimization.

\end{abstract}


\makeatletter
\providecommand{\ceurConference@name}{}
\makeatother

\maketitle

\section{Introduction}
Procedural Content Generation (PCG) automatically creates game content through algorithmic methods often leveraging search or machine learning-based methods (PCGML). Learning-based methods in particular have enabled the automated creation of game content that optimizes designer-specified high-level objectives \cite{todd2023level, nie2025moonshine, khalifa2020pcgrl, earle2024dreamcraft, summerville2018procedural} that would be difficult to express or satisfy through traditional algorithms such as WaveFunctionCollapse (WFC)~\cite{cheng2020automatic}. However, many existing learning-based PCG approaches are primarily applied to visually simplistic domains where the tile set imposes limited adjacency constraints~\cite{todd2023level, nie2025moonshine, khalifa2020pcgrl, earle2024dreamcraft}. We suspect that even methods which already produce satisfactory results in a domain with more aesthetic adjacency constraints like Mario~\cite{lee2023using, wang2022fun, shu2021experience}, would achieve higher robustness and efficiency from offloading the learning of the adjacency rules. We hypothesize that optimization-based PCG methods struggle with complex aesthetic adjacency rules due to the combinatorial explosion of valid tile configurations. To test this hypothesis, we compare traditional approaches that must learn these constraints implicitly against our novel formulation that leverages WFC's constraint propagation mechanism.

While WFC is inherently limited in its ability to optimize global functional properties such as playability, it excels at enforcing fine-grained aesthetic coherence by propagating preferences over possible tile placements via constraint solving \cite{kim2019automatic, karth2021wavefunctioncollapse}. By reformulating WFC as a Markov Decision Process (MDP) and embedding an explicit objective function within this framework, we enable a more generalizable approach to guiding content generation toward designer-specified goals while simultaneously maintaining adherence to aesthetic adjacency constraints.

We optimize this MDP by evolving an action sequence using $\mu + \lambda$ evolution. To understand the implications of the MDP's explicit adjacency constraint guarantees, we compare against evolving the tiles of the final artifact directly, via both naive $\mu + \lambda$ evolution and Feasible Infeasible 2-Population (FI-2Pop) \cite{KIMBROUGH2008310}, where the optimization algorithm must implicitly learn to resolve adjacency violations.

We evaluate these optimization strategies across 3 domains. The binary domain, adapted from \citeauthor{khalifa2020pcgrl}, imposes desired path length constraints as a simplified proxy for functional playability. Notably, varying the target path length allows us to dynamically adjust problem difficulty, enabling a systematic evaluation of how each optimization algorithm scales with increasing task complexity. On the other hand, the biome domains define objective functions that capture global topographic features of distinct biomes, guiding optimization toward semantically coherent and visually pleasing patterns. Finally, we integrate both binary and biome objectives, requiring solutions that jointly optimize metrics pertaining to form and function, thereby reflecting the multi-objective demands typical of real-world game design.

Across all domains and desired path lengths,  non-MDP methods (which incorporate WFC-style constraints into objective functions instead of leveraging them via an MDP) perform noticeably worse; evolving the MDP action sequence leads to faster and more consistent convergence. However, on the most difficult objectives, evolving the action sequence converges very inconsistently, hinting at exploration limitations of $\mu + \lambda$ evolution. 

Concretely, we offer the following contributions:

\begin{itemize}
    \item We demonstrate that forcing learning algorithms to learn local adjacency constraints leads to degraded performance in highly constrained domains.
    \item We present a novel formulation of WFC as an MDP, along with a corresponding \texttt{gym}~\cite{brockman2016openai} environment to facilitate the evaluation of alternative optimization algorithms.
\end{itemize}
\section{Related Work}
\subsection{\textbf{WaveFunctionCollapse Algorithm and Modifications}}

\subsubsection{\textbf{WaveFunctionCollapse}}

WFC is an algorithm for procedural content generation, particularly for tile-based environments. It can create coherent and visually consistent outputs based on simple input examples or constraints~\cite{Gumin_Wave_Function_Collapse_2016}. The algorithm's core functionality resembles constraint satisfaction with a quantum mechanics-inspired approach: maintaining a superposition of possible states for each tile that gradually ``collapses" to definite states through observation and propagation steps~\cite{10.1145/3102071.3110566}.

The algorithm operates in two main modes: the simple tiled model, which uses explicit adjacency rules, and the overlapping model, which automatically extracts patterns from example inputs~\cite{Heaton_2018}. In the both models, WFC can be seen as a form of self-supervised learning that learns a distribution from minimal examples—often just one—and generates similar content by maintaining the learned adjacency patterns. While WFC excels at maintaining local adjacency constraints, it struggles with global optimization objectives that are critical for gameplay, such as ensuring level solvability or balanced resource distribution~\cite{kim2019automatic}.

\subsubsection{\textbf{Enhancements to WaveFunctionCollapse}}
Researchers have developed numerous modifications to improve WFC's capabilities beyond its original formulation. \citeauthor{9421370} provide a comprehensive analysis of WFC as a constraint satisfaction problem, establishing its theoretical foundations and positioning it within the broader context of PCG approaches.

To improve scalability to large environments, \citeauthor{Nie2023ExtendWF} proposed Nested Wave Function Collapse (N-WFC), which decomposes the generation process into nested subproblems. This approach significantly reduces time complexity from exponential to polynomial while maintaining consistency across the generated content. For non-grid environments, researchers have developed graph-based extensions of WFC that enable content generation for arbitrary topologies, enabling applications to 3D worlds and non-uniform structures~\cite{kim2019automatic}.
\citeauthor{LNU_2020} introduce additional extensions to WFC including the path constraint, which enforces global connectivity between specified tiles---addressing one of WFC's key limitations regarding global structure control. This path constraint is distinct from the path-length objective optimized in the present work in that it focuses on paths between designer-specified tiles as opposed to any two most distant tiles on the map, and only ensures the \textit{existence} of such a path, rather than making guarantees about its length. This distinction is significant because optimizing path length requires global reasoning about the entire map structure, which traditional WFC cannot perform. While the path constraint ensures connectivity exists, our work specifically optimizes the longest shortest path between any two points, creating a more challenging optimization problem that must balance local tile placement decisions with their global impact on path topology. \footnote{We note that future work could attempt to optimize path-lengths between specific tiles with a simple modification to our path-length objective function.} \citeauthor{cheng2020automatic} incorporates designer-driven heuristics to steer the generation process toward specific aesthetic or functional outcomes by introducing an automatic rule system along with global, multi‑layer, and distance constraints to provide designers with more control over level layouts \cite{cheng2020automatic}. Yet the approach hinges on enumerated, non-local constraints—tile caps and anchored cells, hard distance windows among entities, and layer-locking of assets—that must be re-engineered for each domain, limiting generality compared with an objective-based formulation that subsumes these preferences in a unified reward.

\subsection{\textbf{Evolutionary Approaches to PCG}}

\subsubsection{\textbf{Evolutionary Algorithms}}
Evolutionary algorithms, widely applied to PCG problems, are a flexible approach to searching the space of possible game content while optimizing for designer-specified objectives~\cite{togelius2011search}. These approaches typically encode content as genomes that evolve through operations such as crossover and mutation, with fitness functions guiding the search toward desired properties.

\subsubsection{\textbf{FI-2Pop for Constrained Optimization}}
The Feasible-Infeasible Two-Population (FI-2Pop) genetic algorithm~\cite{KIMBROUGH2008310}, is a solution to constrained optimization problems that maintains two separate populations: one of feasible solutions that optimize the objective function, and another of infeasible solutions that minimize constraint violations. This approach has proven particularly effective for PCG applications where content must simultaneously satisfy hard constraints (e.g., playability) while optimizing soft objectives (e.g., challenge or balance).

\citeauthor{10.1007/978-3-642-12239-2_14} applied FI-2Pop to procedural level generation, creating game maps that satisfy playability constraints while optimizing for designer preferences. Later, \citeauthor{10.1162/EVCO_a_00123} extended FI-2Pop for constrained novelty search in game level design. This approach uses novelty metrics rather than objective functions to drive search, resulting in a broader exploration of the feasible design space.

\subsection{\textbf{Procedural Content Generation via Reinforcement Learning (PCGRL)}}

\citeauthor{10.5555/3505464.3505478} introduced PCGRL as a novel approach to procedural content generation that frames level design as a sequential decision-making process optimized through reinforcement learning. By formulating level generation as a Markov Decision Process (MDP), PCGRL enables an agent to learn a policy that maximizes expected level quality through incremental modifications. This approach can operate without human-authored examples and generates content extremely quickly once trained.


\section{Problem Domain}
\label{sec:domain}
\begin{figure}
    \centering
    \includegraphics[width=0.34\linewidth]{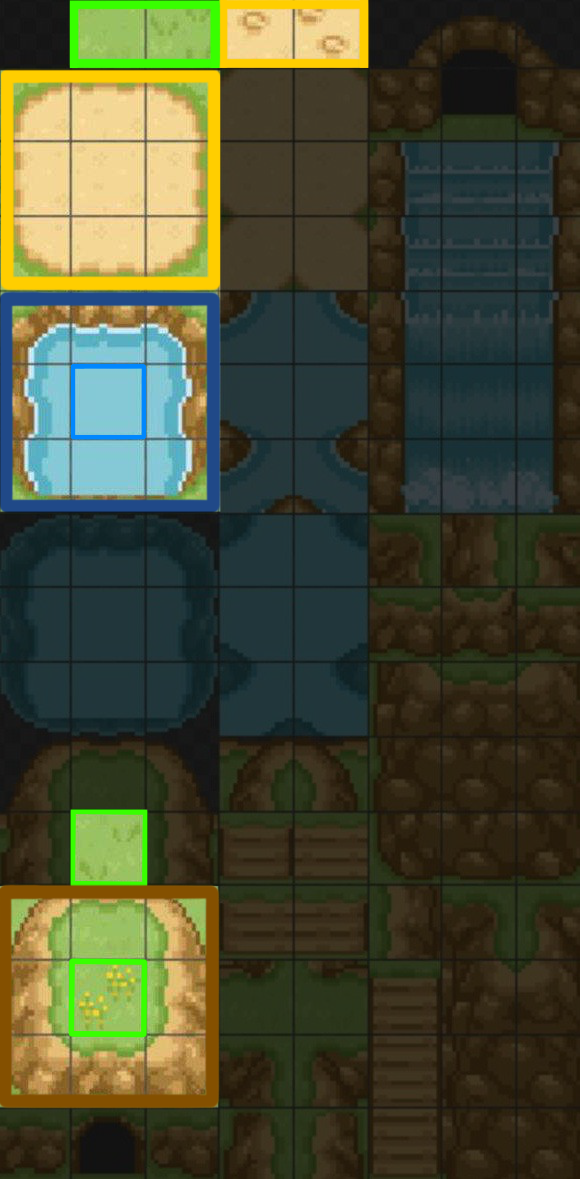}
    \caption{Biome Tileset Pack B - Grassland, Savannah, and Scrubland. Unused tiles are darkened. Path tiles are marked in \textcolor[HTML]{fbd100}{orange}, grass tiles are marked in \textcolor[HTML]{41fe00}{green}, water tiles are marked in \textcolor[HTML]{1a4c8b}{blue}, and hill tiles are marked in \textcolor[HTML]{7c5201}{brown}. The water center tile is marked in \textcolor[HTML]{0b81e7}{light blue}.}
    \label{fig:tiles-sliced}
    \
\end{figure}

All maps and thus objective functions were constructed based on a small subset of \emph{Biome Tileset Pack B - Grassland, Savannah, and Scrubland}~\cite{tileset} which offers combinations of different grass, path, and water tiles. We used a subset of the tiles shown in Figure~\ref{fig:tiles-sliced}. We generate the adjacency rules via manual human labeling, though they could also be extracted from an input image.

\subsection{Binary}
\begin{figure}
    \centering
    \begin{subfigure}[b]{0.30\textwidth}
        \centering
        \includegraphics[width=\textwidth]{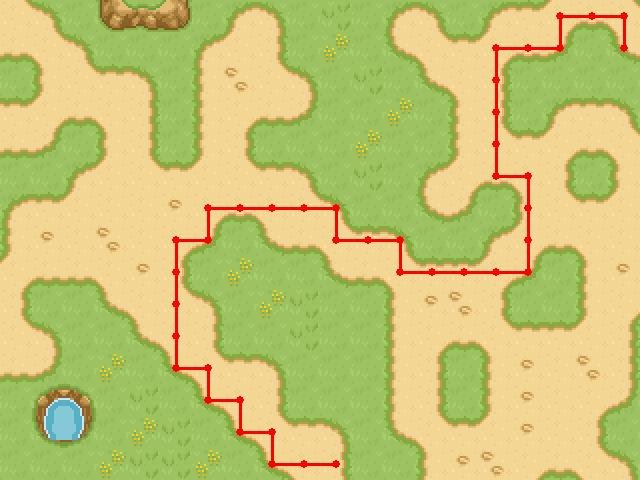}
        \caption{Path Length: 40}
        \label{fig:binary_hard_40}
    \end{subfigure}
    \hfill
    \begin{subfigure}[b]{0.30\textwidth}
        \centering
        \includegraphics[width=\textwidth]{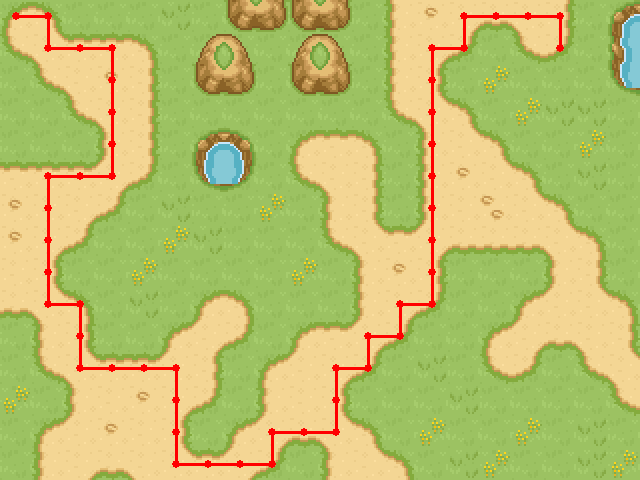}
        \caption{Path Length: 50}
        \label{fig:binary_hard_50}
    \end{subfigure}
    \hfill
    \begin{subfigure}[b]{0.30\textwidth}
        \centering
        \includegraphics[width=\textwidth]{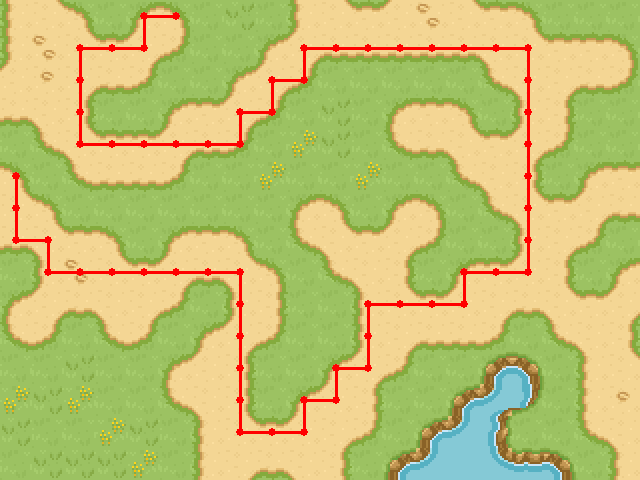}
        \caption{Path Length: 60}
        \label{fig:binary_hard_60}
    \end{subfigure}
        \caption{Optimizing for target path-lengths in the Binary domain. The red line shows the longest shortest path.}
        \label{binary}
\end{figure}

The binary domain (Figure~\ref{binary}), originally introduced in PCGRL~\cite{khalifa2020pcgrl}, tasks the generator with modifying an existing map composed of solid and empty tiles to increase the longest shortest path between any two empty tiles by at least 20 tiles, while ensuring full connectivity of the empty space. In contrast, our formulation requires generating valid maps entirely from scratch that satisfy prescribed path length constraints, rather than modifying preexisting layouts. We also impose a stricter requirement, penalizing deviations from the target by requiring the generated map to achieve a path length of exactly $P$. Formally, for a generated map with longest shortest path length $p$, the objective function awards a score of $-|p - P|$.
While the binary domain in PCGRL comprises a simple ``binary'' tile-set of wall and empty tiles, we use a more sophisticated tileset in which adjacent tiles---e.g. path and grass tiles---can share straight or rounded edges, creating visually smooth transitions between tile-types.

\subsection{Biome/Binary Hybrid}
The hybrid domain adds complexity by dictating the biome surrounding the binary path to better simulate the needs of real game maps. We consider 2 biomes which specify different distributions and arrangements of tiles. All percentage calculations are taken from the tiles of the final artifact, not including those used in binary path calculations. For example, if the objective specifies 50\% water tiles, this means 50\% of the tiles that are not path tiles must be water. We specify the biome objectives below.

\begin{itemize}

\item \textbf{River} ($o_r$).  

Let
\[
  \begin{array}{@{}r@{\quad}p{0.75\columnwidth}@{}}
    r_r   & \text{the number of connected river regions},\\
    \ell  & \text{the length of the current river path},\\
    n_c   & \text{the number of water “center” tiles},\\
    r_l   & \text{the number of connected land regions}.
  \end{array}
\]
We then define:
\begin{equation}
\begin{split}
  o_r = (1 - r_r)
      &+ \min(0,\,\ell - 35) \\
      &- n_c \\
      &+ \min(0,\,3 - r_l)\,.
\end{split}
\end{equation}
The objective attains its maximum value of 0 exactly when all of the following hold:
\begin{itemize}
  \item $r_r = 1$ (exactly one contiguous river region),
  \item $\ell \ge 35$ (river path length of at least 35 tiles),
  \item $n_c = 0$ (no fully surrounded water tiles),
  \item $r_l \le 3$ (no more than three separate land regions).
\end{itemize}
Here, the $(1 - r_r)$ term enforces a single connected channel, the $\min(0,\ell - 35)$ term rewards reaching the target length of 35 tiles, the $-n_c$ penalty encourages a thin, winding river (few interior water tiles), and the cap on $r_l$ discourages excessive fragmentation of the land to prevent it from interrupting the river (Figure~\ref{fig:river}).

\item \textbf{Field} ($o_f$) .  

Let
\[
  \begin{array}{@{}r@{\quad}p{0.75\columnwidth}@{}}
    n_w & \text{the number of water tiles},\\
    n_h & \text{the number of hill tiles},\\
    g   & \text{the percent of grass tiles},\\
    f   & \text{the percent of flower tiles}.
  \end{array}
\]
We then define:
\begin{equation}
\begin{split}
  o_f = -\,n_w - n_h
      &+ \min(0,\,g - 20) \\
      &+ \min(0,\,f - 20)\,.
\end{split}
\end{equation}
The objective attains its maximum value of 0 exactly when all of the following hold:
\begin{itemize}
  \item $n_w = 0$ (no water or shore tiles),
  \item $n_h = 0$ (no hill tiles),
  \item $g \ge 20$ (at least 20\% of tiles are grass),
  \item $f \ge 20$ (at least 20\% of tiles are flowers).
\end{itemize}
Here, the $-n_w$ and $-n_h$ penalties enforce a clear, unbroken grassy field, while the $\min(0,\,g-20)$ and $\min(0,\,f-20)$ terms ensure minimum coverage of grass and flowers. Together, these components encourage the generation of open grasslands with sufficient floral detail (Figure~\ref{fig:grass}).

\begin{figure}
    \centering
    \begin{subfigure}[b]{0.4\textwidth}
        \centering
        \includegraphics[width=\textwidth]{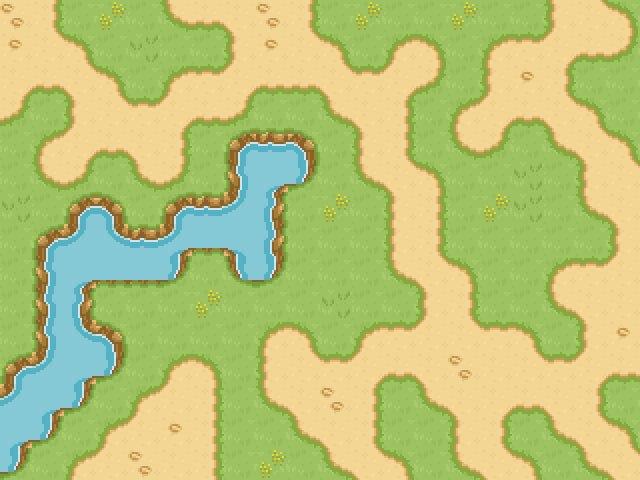}
        \caption{River Biome}
        \label{fig:river}
    \end{subfigure}
    \hfill
    \begin{subfigure}[b]{0.4\textwidth}
        \centering
        \includegraphics[width=\textwidth]{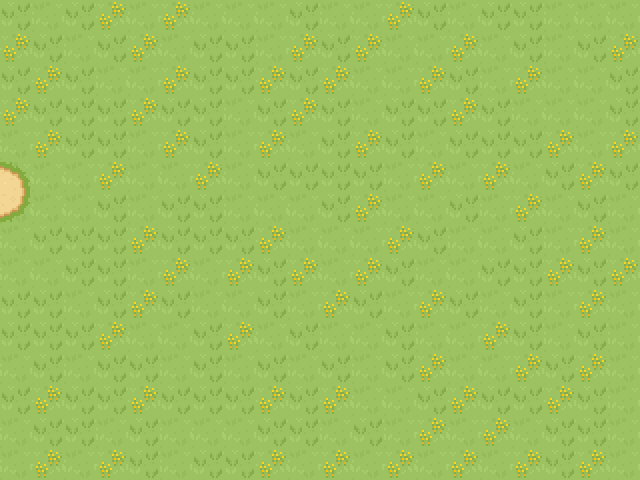}
        \caption{Field Biome}
        \label{fig:grass}
    \end{subfigure}
        \caption{Outputs resulting from the optimization of the Biome objectives}
        \label{fig:biomes}
\end{figure}




\end{itemize}

Since all objective functions have a max reward of 0, we can enforce both binary and biome features by optimizing over the sum of the two objective functions. As such, we get the following objective functions:
\begin{itemize}
    \item hybrid river/binary: $o_{rb}=-|p - P| + o_r$
    \item hybrid field/binary: $o_{fb}=-|p - P| + o_f$.
\end{itemize}

\section{Methods}

All optimization methods have various hyperparameters detailed in Section~\ref{sec:appendix} of the Appendix.
\subsection{Direct Map Evolution}
These methods operate directly on the final artifact and do not leverage WFC. Instead, the optimization process must learn to satisfy the adjacency rules. For a target map of length $\ell$ and width $w$, the genotype is represented as a 2D array of size $\ell \times w$, where each entry contains an integer corresponding to a tile index in the tileset. For example, if position $(x, y)$ contains value $z$, then the $z$th tile is placed at coordinate $(x, y)$ in the artifact, irrespective of whether this placement violates adjacency constraints.

\paragraph{Baseline Evolution.}
The baseline evolutionary algorithm treats each map genotype as an individual and applies standard genetic operators with a penalized fitness that subtracts adjacency violations from raw objective function (Algorithm \ref{alg:naive}). Given objective score $o$ and $v$ adjacency violations, the individuals will receive a fitness of $o-v$. Over $G$ generations it maintains a population of size $N$, selecting the top $\rho N$ individuals by fitness each round.
\begin{algorithm}[!htbp]
\caption{Baseline Evolution}
\label{alg:naive}
\textbf{Input:} generations $G$, population size $N$, survival rate $\rho$\\
\textbf{Output:} Best genome found
\begin{algorithmic}[] 
  \STATE Initialize population $P$ with $N$ random genomes\\
  \STATE Each genome $x$ has an objective score $o_x$, adjacency-constraint violation penalty $v_x$, and fitness $f_x$.
  \FOR{$g = 1$ \TO $G$}
    \FORALL{genome $x \in P$}
      \STATE $(o_x, v_x) \leftarrow \mathrm{Evaluate}(x)$
      \STATE $f_x \leftarrow o_x - v_x$
    \ENDFOR
    \STATE $\text{elites} \leftarrow$ top $\lceil \rho N \rceil$ genomes in $P$ by $f_x$
    \STATE $\text{offspring} \leftarrow \mathrm{Reproduce}(\text{elites},\; N - |\text{elites}|)$
    \STATE $P \leftarrow \text{elites} \cup \text{offspring}$
  \ENDFOR
  \STATE \textbf{return} best feasible genome in $P$
\end{algorithmic}
\end{algorithm}

\paragraph{FI-2Pop.}
FI-2Pop~\cite{KIMBROUGH2008310} attempts to leverage adjacency violations as an exploration medium by maintaining two equal–sized subpopulations, feasible ($F$) and infeasible ($I$), and applies tailored selection criteria to each: objective maximization in $F$ and violation minimization in $I$ (Algorithm~\ref{alg:fi2pop}). Since $I$ is not constrained by the objective function, it is free to explore the boundaries of infeasibility where optimality may lie. The offspring are generated separately to replenish each subpopulation to size $N/2$.

\begin{algorithm}[!htbp]
\caption{FI‑2Pop Evolution}
\label{alg:fi2pop}
\textbf{Input:} generations $G$, population size $N$, survival rate $\rho$\\
\textbf{Output:} Best genome found
\begin{algorithmic}[] 
  \STATE Initialize population $P$ with $N$ random genomes
  \STATE Initialize empty sets $F$ and $I$

  \FOR{$g = 1$ \TO $G$}
      \FORALL{genome $x \in P$}
        \STATE $(o_x, v_x) \leftarrow \mathrm{Evaluate}(x)$
        \IF{$v_x = 0$}
          \STATE $F \leftarrow F \cup \{x\}$
        \ELSE
          \STATE $I \leftarrow I \cup \{x\}$
        \ENDIF
      \ENDFOR
    \STATE $F_s \leftarrow$ top $\lceil \rho\,|F| \rceil$ genomes in $F$ by objective score $(o_x)$
    \STATE $I_s \leftarrow$ top $\lceil \rho\,|I| \rceil$ genomes in $I$ by lowest violation $(v_x)$
    \STATE $O_F \leftarrow \mathrm{Reproduce}(F_s,\; \tfrac{N}{2} - |F_s|)$
    \STATE $O_I \leftarrow \mathrm{Reproduce}(I_s,\; \tfrac{N}{2} - |I_s|)$
    \STATE $P \leftarrow O_F \cup O_I$
  \ENDFOR
  \STATE \textbf{return} best genome in $F$
\end{algorithmic}
\end{algorithm}

\subsection{MDP Representation.}
By formalizing WFC as a Markov Decision Process (MDP), we leverage its guarantees to offload the burden of learning adjacency constraints from the optimizer. This reformulation transforms the generation problem into a sequential decision process where every action results in a valid intermediate configuration.

\begin{algorithm}[!htbp]
\caption{WFC-MDP: One Environment Step}
\label{alg:wfc_mdp_step}
\textbf{Input:} belief grid $G \in \{0,1\}^{\ell \times w \times n_t}$, adjacency rules $A$, agent action (tile logits) $a$\\
\textbf{Output:} updated grid $G'$, reward $r$
\begin{algorithmic}[]
  \STATE $(x^\star, y^\star) \leftarrow \mathrm{FindLowestEntropyCell}(G)$
  \STATE $m \leftarrow G[y^\star, x^\star, :]$ \hfill\textit{feasibility mask at selected cell}
  \IF{$(x^\star, y^\star) = \varnothing \text{ or } \sum_{t=1}^{T} m_t = 0$}
    \STATE $r \leftarrow -1000$
  \ENDIF
  \STATE $p \leftarrow \text{softmax}(a)$
  \FORALL{t $ \in \{1,\dots,n_t\}$}
    \IF{$m[t] = 0$}
        \STATE $p[t] \leftarrow 0$
    \ENDIF
  \ENDFOR
  \STATE $t^\star \leftarrow \arg\max p$
  \STATE $G \leftarrow \mathrm{Collapse}(G,\ (x^\star, y^\star) \gets t^\star)$
  \STATE $(G,\ \texttt{ok}) \leftarrow \mathrm{PropagateConstraints}(G,\ A,\ (x^\star, y^\star))$
  \IF{\textbf{not} $\texttt{ok}$}
    \STATE $r \leftarrow -1000$
  \ELSIF{$\mathrm{IsFullyCollapsed}(G)$}
    \STATE $r \leftarrow \text{Objective}(G)$
  \ENDIF
  \STATE \textbf{return} $(G,\ r)$
\end{algorithmic}
\end{algorithm}

\subsubsection{State Representation}
We define each state $s_t$ as the current configuration of the WFC grid at timestep $t$. For a target map of size $\ell \times w$ with $n_t$ tile types, $s_t$ is represented by an $\ell \times w \times n_t$ binary tensor $G_t$, where each depth slice encodes the feasibility of a tile at a cell:
\begin{itemize}
  \item $G_t[x,y,i] = 1$ indicates that tile type $i$ is currently \emph{feasible} at cell $(x,y)$.
  \item $G_t[x,y,i] = 0$ indicates that tile type $i$ is \emph{infeasible} at cell $(x,y)$.
\end{itemize}
A cell is \emph{collapsed} if and only if it has a one-hot feasibility vector ($\sum_{i=1}^{n_t} G_t[x,y,i] = 1$); otherwise it is \emph{uncollapsed} ($\sum_{i=1}^{n_t} G_t[x,y,i] > 1$). The MDP terminates when every cell is collapsed (equivalently, when all $\ell \times w$ cells have one-hot feasibility vectors), yielding the final artifact.

\subsubsection{Action Representation}
At each timestep $t$, the action $a_t$ specifies which tile to collapse in a single uncollapsed cell (Algorithm~\ref{alg:wfc_mdp_step}). It is parameterized as an $n_t$-dimensional logit vector, with each entry corresponding to a possible tile; these logits lead to corresponding per tile probability after softmax. To ensure compliance with adjacency constraints, we mask out invalid tiles by setting their probabilities to zero. The selected action is then $\arg\max$ over the probabilities, ensuring that each collapse step is constraint-respecting. Since exactly one tile is collapsed per action, a complete map requires a sequence of $\ell \times w$ actions.

\subsubsection{Objective Structure}
We adopt a sparse objective model: intermediate states contribute no score, and only the terminal state is evaluated using the objectives defined in Section~\ref{sec:domain}. If WFC enters a contradiction where a cell has no valid tiles remaining, the process truncates immediately and incurs a large negative objective of -1000, thus discouraging invalid map constructions.

\subsection{Evolving an Action Sequence}

We use a standard $\mu + \lambda$ evolutionary algorithm to optimize the full sequence of WFC collapse actions (Algorithm~\ref{alg:action_sequence_evolution}). Each individual in the population encodes a fixed-length sequence of collapse decisions, represented as logits over the tile set at each of the $\ell \times w$ positions. Two genotype shapes are supported:

\begin{itemize}
  \item \textbf{1D representation:} a flat sequence of length $\ell \times w$ such that element at position $x$ is the action applied at time step $x$.
  \item \textbf{2D representation:} a grid-aligned sequence of dimensions $\ell \times w$, where each action's position is inferred from WFC's next-collapse coordinates. Action at genotype coordinate $(x,y)$ corresponds to collapsing the tile at $(x,y)$. 
\end{itemize}

\begin{algorithm}[!htbp]
\caption{Evolving an Action Sequence}
\label{alg:action_sequence_evolution}
\textbf{Input:} generations $G$, population size $N$, survival rate $\rho_s$\\
\textbf{Output:} Best genome found
\begin{algorithmic}[]
  \STATE Initialize population $P$ with $N$ action sequences
  \FOR{$g = 1$ \TO $G$}
    \FORALL{genome $x \in P$}
      \STATE $s \leftarrow s_0$; $O_x \leftarrow 0$; $v_x \leftarrow 0$
      \FOR{$t = 1$ \TO $\ell \times w$}
        \STATE $a_t \leftarrow x[t]$
        \STATE $(s', r_t) \leftarrow \mathrm{env.step}(a_t)$
        \STATE $O_x \leftarrow O_x + r_t$
        \STATE $s \leftarrow s'$
      \ENDFOR
    \ENDFOR
    \STATE $\text{elites} \leftarrow$ top $\lceil \rho_s N \rceil$ genomes in $P$ by $f_x$
    \STATE $\text{offspring} \leftarrow \mathrm{Reproduce}(\text{elites},\; N - |\text{elites}|)$
    \STATE Initialize population $R$ with $(N - |\text{elites}|)$ action sequences
    \STATE $P \leftarrow \text{elites} \cup \text{offspring}$
  \ENDFOR
  \STATE \textbf{return} best feasible genome in $P$
\end{algorithmic}
\end{algorithm}

\section{Results}
\label{sec:results}

We evaluate each optimization method across desired path lengths 10-100 in intervals of 10 for both binary and hybrid domains. Convergence robustness is defined as the proportion of runs that achieve the maximal reward of 0, while sample efficiency is measured by the number of generations it takes to evolve at least one population member with reward of 0. All methods were run with a fixed sample budget (population size = 48) to enable fair comparisons.

Table~\ref{tab:binary_convergence} reports performance across increasing path length targets in the binary domain. As expected, longer path lengths correspond to increased difficulty as all methods show reduced convergence rates and increased generations to convergence with rising desired path length $P$.

However, even at low difficulty ($P \leq 40$), non-MDP baselines occasionally fail to converge. In contrast, both MDP-based evolution strategies exhibit perfect or near-perfect convergence and require fewer generations. This early divergence suggests that directly modeling constraint satisfaction via WFC confers immediate robustness advantages, even in seemingly simple settings.

As $P$ increases, the performance gap widens. At $P=60$, non-MDP methods converge in less than 20\% of runs and exhibit inefficiency when they do, with baseline requiring over 170 generations on average. Evolution 1D and 2D remain below 60 generations and achieve convergence in 84\% and 72\% of runs respectively, underscoring their superior scalability.

For $P \geq 70$, all non-MDP methods fail almost entirely. MDP methods, while less consistent, still achieve convergence in up to 35\% of runs (1D) and 15\% (2D). Interestingly, the baseline was able to converge once, in relatively few generations. This outlier suggests a rare, favorable initialization or fortuitous stochasticity early on in the evolutionary process rather than systematic optimization success.

In the hybrid river/binary domain (Table~\ref{tab:river_convergence}), MDP-based methods again demonstrate clear robustness advantages. For moderate difficulty ($20 \leq P \leq 40$), 1D and 2D evolution achieve convergence in at least 36\% and 45\% of runs respectively, requiring roughly 50–100 generations on average. In contrast, baseline and fi2pop converge in at most 14\% and 4\% of runs, and when they do succeed, require over 80–120 generations. At $P = 50$, both MDP methods maintain non-trivial success rates ($\geq$19\%) with convergence typically within 80–100 generations, whereas baseline converges only 4\% of the time and fi2pop fails entirely. Beyond this threshold, only MDP methods achieve any convergence at all: both 1D and 2D occasionally succeed for $P \geq 60$, though with growing variance in required generations, while non-MDP methods collapse to zero convergence. The performance gap is more stark in Table~\ref{tab:grass_convergence} and the hybrid field/binary domain. Non-MPD methods completely fail to converge across almost all target path lengths.

\begin{table}[ht]
\centering
\begin{adjustbox}{max width=\textwidth}
\begin{tabular}{llllllllll}
\toprule
Method & Metric & 10 & 20 & 30 & 40 & 50 & 60 & 70 & 80 \\
\midrule
\multirow[t]{2}{*}{evo 1d} & Generations & \textbf{5.9±0.6} & 8.4±0.6 & 6.7±0.5 & 9.4±0.6 & \textbf{21.9±1.4} & 54.8±3.2 & \textbf{114.7±10.4} & 228.0±73.0 \\
 & Converged\% & \textbf{1.00} & \textbf{1.00} & \textbf{1.00} & \textbf{1.00} & \textbf{0.97} & \textbf{0.84} & \textbf{0.35} & \textbf{0.02} \\
\midrule
\multirow[t]{2}{*}{evo 2d} & Generations & 7.4±0.6 & \textbf{6.2±0.5} & \textbf{5.8±0.5} & \textbf{9.1±0.8} & 23.6±1.5 & \textbf{41.9±2.8} & 74.9±13.4 & \textbf{136.0±74.0} \\
 & Converged\% & \textbf{1.00} & \textbf{1.00} & \textbf{1.00} & 0.98 & 0.95 & 0.72 & 0.15 & \textbf{0.02} \\
\midrule
\multirow[t]{2}{*}{baseline} & Generations & 14.8±2.5 & 11.0±1.8 & 8.8±0.9 & 22.8±3.2 & 70.2±12.2 & 171.8±48.9 & 35 & — \\
 & Converged\% & 0.85 & 0.95 & 1.00 & 0.95 & 0.65 & 0.16 & 0.01 & — \\
\midrule
\multirow[t]{2}{*}{fi2pop} & Generations & 9.7±1.3 & 12.9±1.5 & 8.4±1.3 & 17.8±2.2 & 30.3±4.3 & 60.2±14.2 & — & — \\
 & Converged\% & 0.96 & 0.91 & 0.92 & 0.84 & 0.49 & 0.08 & — & — \\
\bottomrule
\end{tabular}
\end{adjustbox}
\caption{Convergence performance across increasing binary path-length objectives ($P$). Each method is evaluated on convergence robustness (proportion of runs achieving optimal reward) and sample efficiency (mean generations to convergence with standard error). Results highlight the superior scalability of MDP-based approaches. Bold values indicate the highest convergence proportion and the lowest statistically significant mean generations per target path length. A visual representation can be found in the Appendix (Figure~\ref{fig:binary_convergence}).}
\label{tab:binary_convergence}
\end{table}

\begin{table}
\begin{adjustbox}{max width=\textwidth}
\begin{tabular}{lllllllllll}
\toprule
Method & Metric & 10 & 20 & 30 & 40 & 50 & 60 & 70  \\
\midrule
\multirow[t]{2}{*}{evo 1d} & Generations & 62.0±6.5 & 59.7±6.0 & 77.3±5.7 & 98.5±10.4 & 78.7±9.5 & \textbf{78.8±26.9} & — \\
 & Converged\% & \textbf{0.42} & \textbf{0.45} & \textbf{0.59} & \textbf{0.46} & \textbf{0.28} & \textbf{0.05} & — \\
\cline{1-9}
\multirow[t]{2}{*}{evo 2d} & Generations & \textbf{38.3±9.8} & 49.7±6.0 & \textbf{52.2±4.7} & \textbf{51.2±6.3} & \textbf{48.9±8.0} & 92.0±37.0 & \textbf{36} \\
 & Converged\% & 0.11 & \textbf{0.45} & 0.42 & 0.36 & 0.19 & 0.03 & \textbf{0.01} \\
\cline{1-9}
\multirow[t]{2}{*}{baseline} & Generations & 98.5±11.5 & 89.6±18.7 & 114.6±25.9 & 111.7±22.3 & 70.7±13.0 & — & — \\
 & Converged\% & 0.03 & 0.09 & 0.12 & 0.14 & 0.04 & — & —\\
\cline{1-9}
\multirow[t]{2}{*}{fi2pop} & Generations & 40.5±39.5 & 48 & 136 & 82.3±40.8 & —  & — & — \\
 & Converged\% & 0.03 & 0.01 & 0.01 & 0.04 & — & — & —   \\
\bottomrule
\end{tabular}
\end{adjustbox}
\caption{Convergence performance across increasing binary path-length objectives under the hybrid river/binary domain ($P$). Each method is evaluated on convergence robustness (proportion of runs achieving optimal reward) and sample efficiency (mean generations to convergence with standard error). Almost all experiments converging on less than 50\% of runs, MDP base methods maintain a noticeable lead in both metrics. Bold values indicate the highest convergence proportion and the lowest statistically significant mean generations per target path length. A visual representation can be found in the Appendix (Figure~\ref{fig:river_convergence}).}
\label{tab:river_convergence}
\end{table}

\begin{table}
\begin{adjustbox}{max width=\textwidth}
\begin{tabular}{llllllll}
\toprule
Method & Metric & 10 & 20 & 30 & 40 & 50 & 60 \\
\midrule
\multirow[t]{2}{*}{evo 1d} & Generations & 66.8±13.9 & \textbf{80.2±11.7} & 155.1±24.4 & 344.4±37.6 & 551.7±104.2 & 472.0±214.0 \\
 & Converged\%  & \textbf{0.47} & \textbf{0.80} & \textbf{0.90} & \textbf{0.62} & \textbf{0.15} & \textbf{0.05}\\
\cline{1-8}
\multirow[t]{2}{*}{evo 2d} & Generations  & 59.5±9.1 & 94.9±15.0 & \textbf{148.1±19.9} & \textbf{189.6±46.2} & \textbf{260} & \textbf{160.0±159.0} \\
 & Converged\%  & 0.35 & 0.39 & 0.30 & 0.15 & 0.01 & 0.03\\
\cline{1-8}
\multirow[t]{2}{*}{baseline} & Generations  & — & — & — & — & — & — \\
 & Converged\%  & — & — & — & — & — & — \\
\cline{1-8}
\multirow[t]{2}{*}{fi2pop} & Generations  & 27.0±10.5 & — & — & — & — & — \\
 & Converged\%  & 0.04 & — & — & — & — & — \\
\bottomrule
\end{tabular}
\end{adjustbox}
\caption{Convergence performance across increasing binary path-length objectives  under hybrid the field/binary domain ($P$). Each method is evaluated on convergence robustness (proportion of runs achieving optimal reward) and sample efficiency (mean generations to convergence with standard error). Results highlight the superior scalability of MDP-based approaches, especially at higher difficulty levels. Bold values indicate the highest convergence proportion and the lowest statistically significant mean generations per target path length. A visual representation can be found in the Appendix (Figure~\ref{fig:grass_convergence}).}
\label{tab:grass_convergence}
\end{table}

\subsection{Implications and Cross-Domain Insights}

Two key patterns emerge:

\begin{enumerate}
    \item \textbf{MDP Encapsulation of Constraints is Crucial}: Across all desired path lengths, methods that offload constraint enforcement to WFC consistently outperform those that must learn it implicitly. This discrepancy is especially pronounced given more difficult objectives (i.e. higher target path lengths, hybrid biome/binary domains), where the feasibility space is severely constrained.
    \item \textbf{Feasible Region Shrinkage Limits Optimization}: At high path lengths, even MDP methods fail to converge reliably. This likely stems from the exponentially shrinking volume of the feasible space and limited tendency toward exploration in vanilla $\mu+\lambda$ evolution---as compared to e.g. Quality Diversity~\cite{pugh2016quality}. Despite valid intermediate states, the reward landscape remains highly sparse and multi-modal.
\end{enumerate}

These findings highlight a fundamental insight: procedural generation under complex constraints benefits most when constraint satisfaction is externalized and search is guided through structurally aligned representations. The clear failure of joint optimization approaches, particularly in aesthetically constrained domains, emphasizes the importance of architectural modularity in generative design systems.

\section{Discussion}

\subsection{Other Biomes}
\begin{figure}
    \centering
    \begin{subfigure}[b]{0.29\textwidth}
        \centering
        \includegraphics[width=\textwidth]{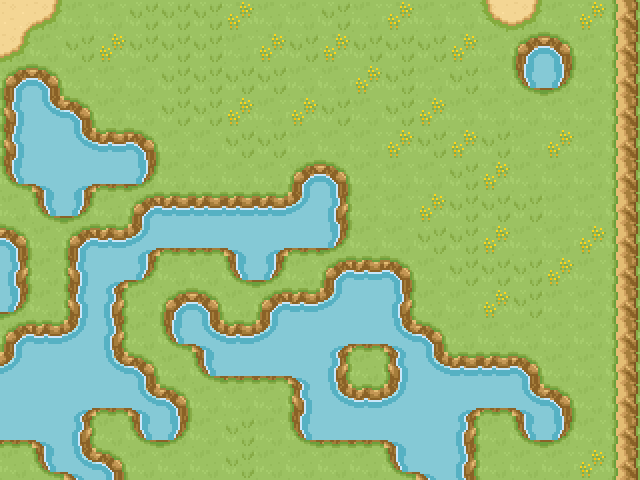}
        \caption{Pond Biome}
        \label{fig:pond}
    \end{subfigure}
    \hfill
    \begin{subfigure}[b]{0.29\textwidth}
        \centering
        \includegraphics[width=\textwidth]{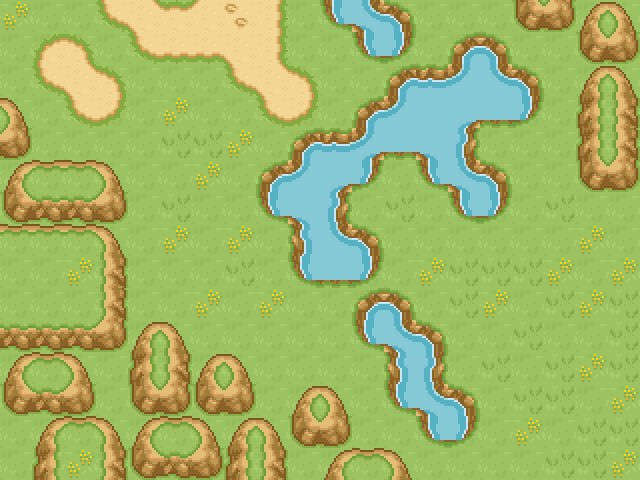}
        \caption{Pond and Hill Biome}
        \label{fig:ponds_and_hills}
    \end{subfigure}
    \hfill
    \begin{subfigure}[b]{0.29\textwidth}
        \centering
        \includegraphics[width=\textwidth]{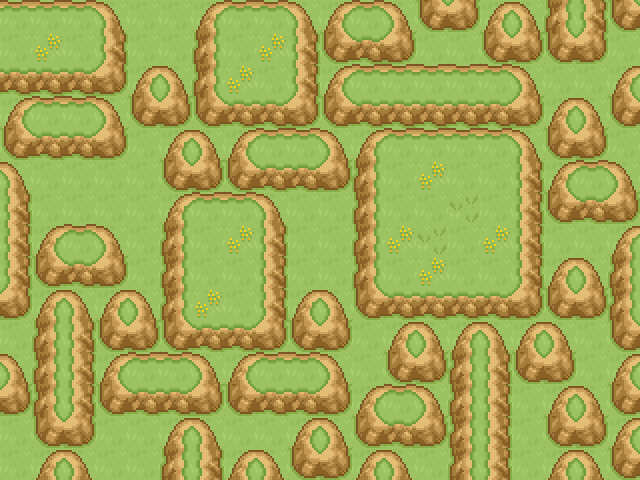}
        \caption{Hill Biome}
        \label{fig:hill}
    \end{subfigure}
        \caption{Outputs resulting from the optimization of other Biome objectives}
        \label{fig:other_biomes}
\end{figure}

WFC-MDP is compatible with various domains without necessitating the bespoke engineering required by more traditional extensions to WFC. By changing the objective function, WFC-MDP is able to optimize for other gameplay artifacts like pond and hill biomes (Figure~\ref{fig:other_biomes}). These biomes were excluded from the main results as they were not adapted to work in conjunction with binary either because they proved excessively difficult to optimize or were not structured to allow long continuous paths; instead they illustrate the flexibility of WFC-MDP. 
We leave the optimization of these more challenging hybrid objectives to future work that builds on our WFC-MDP formulation.

\subsection{Representation Matters}
We evaluated two encoding schemes for collapse sequences: a 2D representation that maps actions to fixed spatial locations, and a 1D representation that encodes a strict sequential order. The 2D encoding reduces the risk of early mutations cascading across the map by localizing genetic variation, resulting in smoother and more stable optimization dynamics. Conversely, the 1D encoding allows mutations in earlier steps to significantly influence subsequent tile collapses, which introduces instability but encourages broader exploration. Empirically, we observe that 1D encodings tend to converge more frequently possibly due to this exploratory behavior. In contrast, 2D encodings exhibit better sample efficiency, aligning with their more localized impact. These trade-offs suggest promising directions for future research, such as hybrid encoding schemes that balance exploration and stability.

\subsection{Toward Scalable and Interactive Generation}
Our method operates on consumer hardware, making it viable for map generation during the game development process. However, as problems get more complex, live evolution becomes time consuming. Further research can explore learning generalized policies that are capable of generating multiple artifacts from single training instance. For example, one could train a Reinforcement Learning policy to output/edit the tile-type logits in our 1D or 2D evolutionary genomes given varying target path-lengths as in \cite{earle2021learning}, with reward equal to the score of a map collapsed via WFC over these logits. Alternatively, one could apply Imitation Learning to the data generated by the evolutionary processes in this paper, similar to \cite{khalifa2022mutation}, and/or use such Imitation Learning to jump-start the RL process outlined above. 
Future work could explore additional heuristics beyond path length optimization, such as requirements for the level to contain important items (e.g., treasures, keys, or strategic resources) or specific gameplay mechanics (e.g., cover points, chokepoints). These would provide additional functional objectives to test the generality of the WFC-MDP approach. Additionally, RL extensions could leverage the \texttt{gym} environment to test alternative optimization methods on the WFC-MDP formulation.

\section{Limitations}

\subsection{Optimization Parameters}
We used a fixed population size of 48 across all algorithms due to hardware constraints and to maintain a consistent measure of sample efficiency. This choice may not reflect the optimal configuration for each method. In particular, FI‑2Pop’s dual‑population architecture may be disproportionately affected by smaller population sizes.

\subsection{Observation Utilization}
In an MDP formulation, the agent’s observation can inform optimal action selection. Agents can make more optimal tile selections if given the partially collapsed map and the next collapse position. However, our evolution-based implementations operate over the full action sequence in advance and do not leverage intermediate observations. 
By instead evolving e.g. a neural network controller to output tile logits at each step of the WFC-MDP, these observations could be leveraged to improve performance.

\subsection{Optimization Limitations}
Although MDP-based methods demonstrate improved consistency and sample efficiency on challenging tasks, the standard $\mu+\lambda$ evolutionary strategy often exhibits inadequate exploration, leading to poor convergence on the hardest problems. We hypothesize that the large PCG state space and highly multimodal fitness landscape demand algorithms with stronger exploration capabilities, such as Quality Diversity evolutionary algorithms \cite{pugh2016quality} or Novelty Search \cite{lehman2011evolving}. 

\section{Conclusion}
This work recasts WaveFunctionCollapse (WFC) as a Markov Decision Process (MDP), enabling optimizers to sidestep the combinatorial burden of learning tile adjacency rules by offloading constraint enforcement to WFC’s propagation mechanism. By evaluating this formulation across various scalable and progressively constrained domains, we uncover a central insight: explicit algorithmic decoupling of constraint satisfaction from objective optimization dramatically improves both convergence reliability and sample efficiency.

Our WFC-MDP framework not only succeeds where traditional methods collapse, but also reveals how reframing generation as a sequence of valid state transitions exposes a richer interface for guidance and learning. The domain’s scalability allowed us to observe how even strong inductive biases (like constraint propagation) falter without sufficient exploration capacity, suggesting fertile ground for hybrid learning approaches.

Looking ahead, our formulation opens the door to more controllable and expressive uses of WFC. Because objective functions are more generalizable than hard-coded global constraints, they provide a natural bridge to machine learning methods, particularly those that rely on flexible reward signals or policy learning. By enabling WFC to support objective driven control over layout generation, this work establishes a blueprint for scalable, constraint-aware, and ML-compatible PCG systems.

\section{Appendix}
\label{sec:appendix}

\subsection{Final Hyperparameter Settings}
Table~\ref{tab:hparams_combined} records our tuned hyperparameters.

\begin{table}[ht]
\centering
\setlength{\tabcolsep}{6pt}
\renewcommand{\arraystretch}{1.15}
\begin{adjustbox}{max width=\textwidth}
\begin{tabular}{llcccc}
\toprule
& & \multicolumn{4}{c}{\textbf{Optimization method}} \\
\cmidrule(lr){3-6}
\textbf{Domain} & \textbf{Parameter} & \textbf{Baseline} & \textbf{FI--2Pop} & \textbf{Action Seq (1D)} & \textbf{Action Seq (2D)} \\
\midrule

\multirow{6}{*}{Binary}
 & \texttt{number\_of\_actions\_mutated\_mean}               & 89        & 162       & 97        & 44 \\
 & \texttt{number\_of\_actions\_mutated\_standard\_deviation} & 157.2498  & 196.1993  & 120.0876  & 28.2708 \\
 & \texttt{action\_noise\_standard\_deviation}               & 0.0810    & 0.0418    & 0.1296    & 0.1409 \\
 & \texttt{survival\_rate}                                   & 0.5211    & 0.3552    & 0.4151    & 0.2328 \\
 & \texttt{cross\_over\_method}                              & 1 (ONE\_POINT) & 1 (ONE\_POINT) & 1 (ONE\_POINT) & 0 (UNIFORM) \\
 & \texttt{cross\_or\_mutate}                                & 0.8324    & 0.9871    & 0.7453    & 0.9557 \\
\midrule

\multirow{6}{*}{Hybrid River/Binary}
 & \texttt{number\_of\_actions\_mutated\_mean}               & 1         & 142       & 79        & 48 \\
 & \texttt{number\_of\_actions\_mutated\_standard\_deviation} & 56.7544   & 69.3442   & 111.7231  & 14.0969 \\
 & \texttt{action\_noise\_standard\_deviation}               & 0.1452    & 0.0413    & 0.1087    & 0.0647 \\
 & \texttt{survival\_rate}                                   & 0.7988    & 0.4726    & 0.4133    & 0.3077 \\
 & \texttt{cross\_over\_method}                              & 1 (ONE\_POINT) & 1 (ONE\_POINT) & 1 (ONE\_POINT) & 0 (UNIFORM) \\
 & \texttt{cross\_or\_mutate}                                & 0.9633    & 0.9130    & 0.9930    & 0.8902 \\
\midrule

\multirow{6}{*}{Hybrid Field/Binary}
 & \texttt{number\_of\_actions\_mutated\_mean}               & 86        & 178       & 23        & 132 \\
 & \texttt{number\_of\_actions\_mutated\_standard\_deviation} & 146.9724  & 68.7875   & 0.5458    & 56.0667 \\
 & \texttt{action\_noise\_standard\_deviation}               & 0.3916    & 0.0125    & 0.4937    & 0.0407 \\
 & \texttt{survival\_rate}                                   & 0.7513    & 0.7720    & 0.1861    & 0.3245 \\
 & \texttt{cross\_over\_method}                              & 0 (UNIFORM) & 0 (UNIFORM) & 1 (ONE\_POINT) & 1 (ONE\_POINT) \\
 & \texttt{cross\_or\_mutate}                                & 0.6876    & 0.7570    & 0.8241    & 0.8415 \\
\bottomrule
\end{tabular}
\end{adjustbox}
\caption{We employed Optuna \cite{akiba2019optuna} to automatically tune evolutionary hyperparameters for each of our three methods and across all experiments. For each method/experiment we ran 20 trials, optimizing the cumulative best reward after 20 training attempts with 100 generations each.}
\label{tab:hparams_combined}
\end{table}

\subsubsection{Hyperparameter Definitions}
\begin{description}
  \item[\texttt{number\_of\_actions\_mutated\_mean}] (int)  
    Expected number of genome actions to mutate each generation.
  \item[\texttt{number\_of\_actions\_mutated\_standard\_deviation}] (float)  
    Standard deviation for truncated‐normal sampling of the mutation count.
  \item[\texttt{action\_noise\_standard\_deviation}] (float)  
    Standard deviation of Gaussian noise added to each mutated action’s value.
  \item[\texttt{survival\_rate}] (float)  
    Fraction of the population preserved into the next generation.
  \item[\texttt{cross\_over\_method}] (enum)  
    Crossover strategy:
    \begin{description}
      \item[0 = UNIFORM] Gene‐wise mixing: each child gene is chosen from one parent with 50\% probability.
      \item[1 = ONE\_POINT] Single cut‐point crossover: parents swap tails at one random index.
    \end{description}
  \item[\texttt{cross\_or\_mutate\_proportion}] (float)  
    Fraction of offspring produced via crossover.
\end{description}

\subsection{Convergence Behavior Plots}

\begin{figure}[!h]
    \centering
    \begin{subfigure}[b]{0.33\textwidth}
        \centering
        \includegraphics[width=\textwidth]{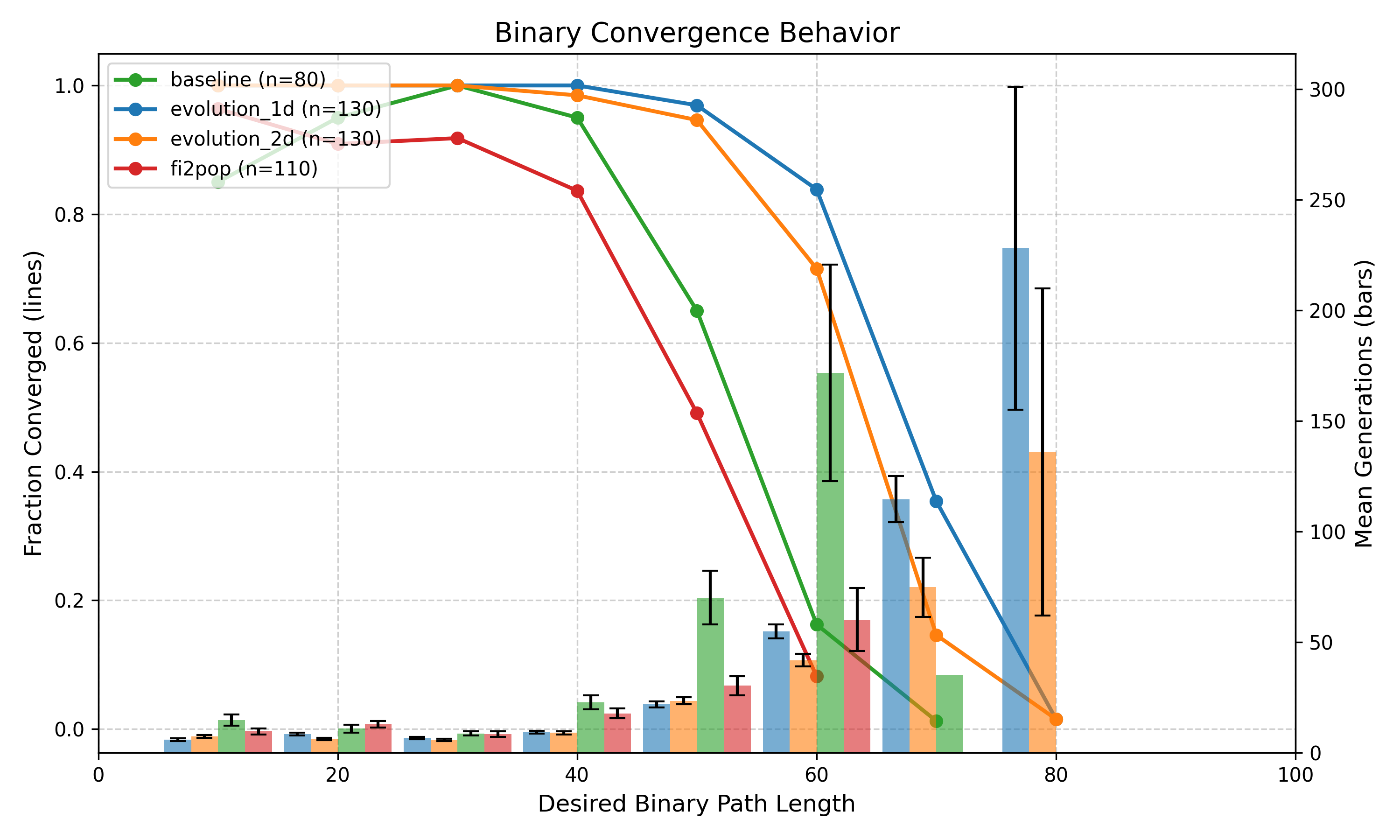}
        \caption{Plot of Table \ref{tab:binary_convergence}}
        \label{fig:binary_convergence}
    \end{subfigure}
    \hfill
    \begin{subfigure}[b]{0.33\textwidth}
        \centering
        \includegraphics[width=\textwidth]{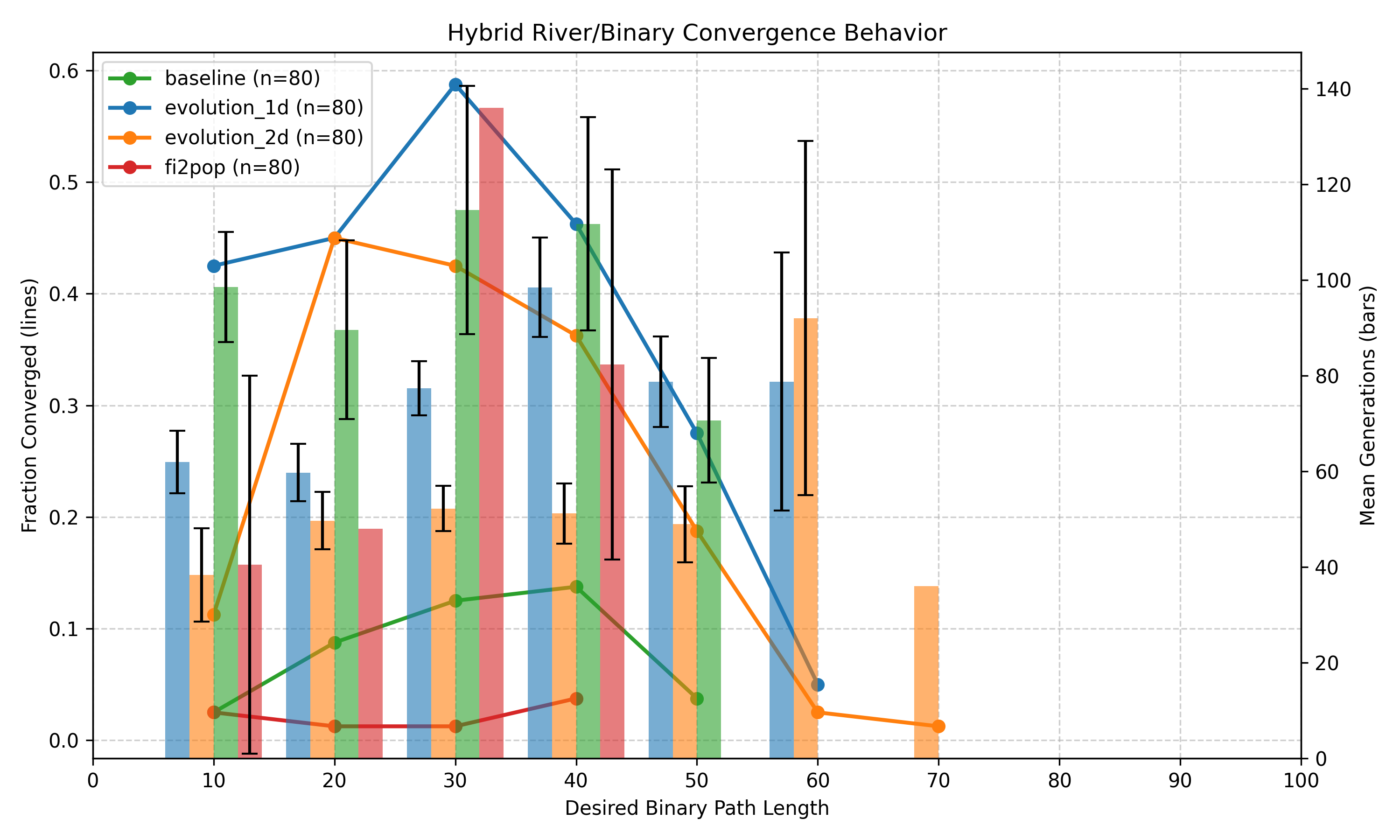}
        \caption{Plot of Table \ref{tab:river_convergence}} 
    \label{fig:river_convergence}
    \end{subfigure}
    \hfill
    \begin{subfigure}[b]{0.33\textwidth}
        \centering
        \includegraphics[width=\textwidth]{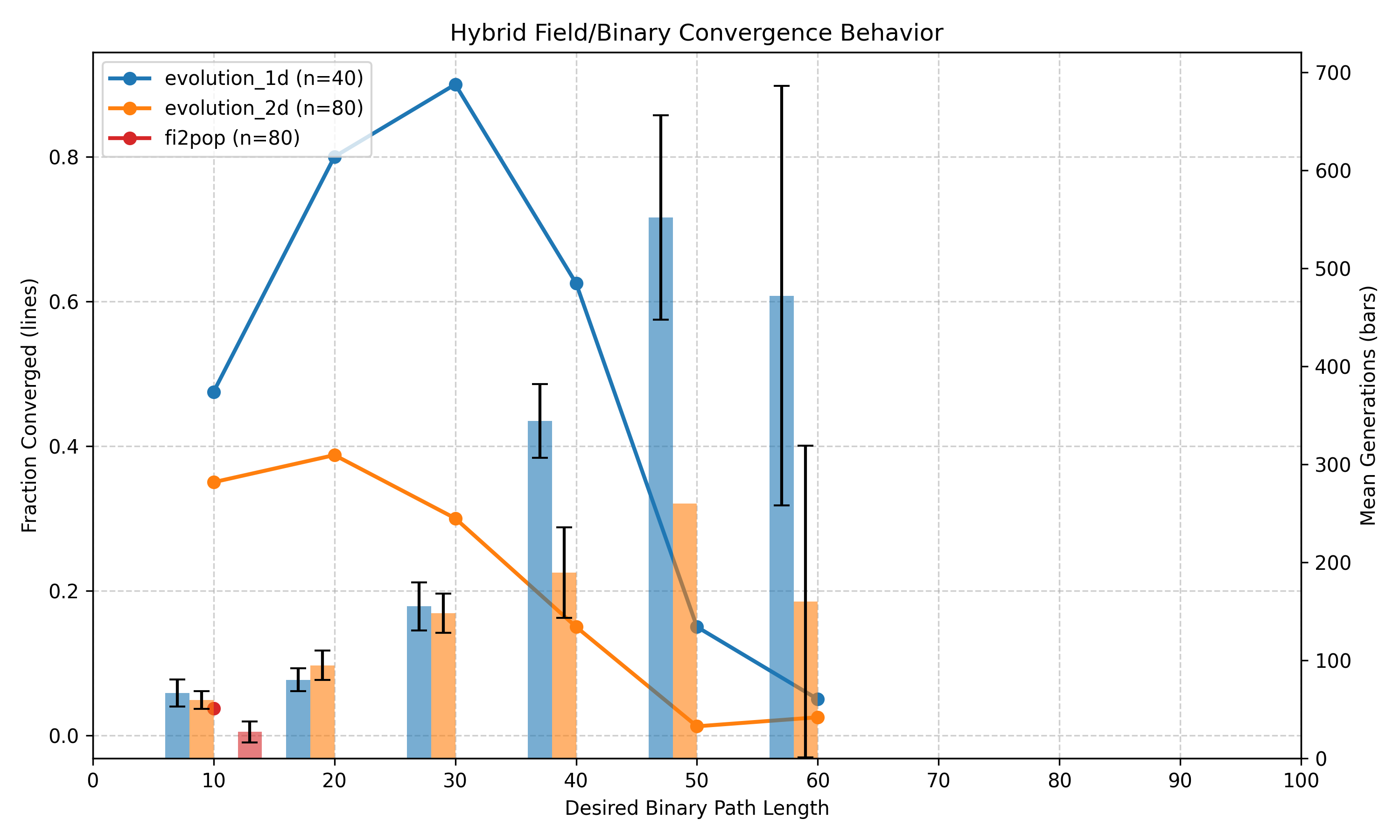}
        \caption{Plot of Table \ref{tab:grass_convergence}} 
        \label{fig:grass_convergence}
    \end{subfigure}
        \caption{Plot which serve as a visual representation of the convergence behavior expressed in a corresponding tables. The lines correlate with the fraction of converged training samples (left y axis — higher is better) and the bars correlate to the mean generations to successfully converge  (right y axis — lower is better).}
        \label{fig:Convergence Data}
\end{figure}

Plots correlated with the tables in Section~\ref{sec:results} (Figure~\ref{fig:Convergence Data}).

\bibliography{sample-ceur}

\end{document}